\documentclass[letterpaper, 10 pt, conference]{ieeeconf}  
\newcommand{\setting}{\textbf{\textsc{BEVMapMatch}}}
\usepackage{url}
\usepackage{graphicx}
\usepackage{multirow}
\usepackage{booktabs}
\usepackage{adjustbox}
\usepackage{hyperref}
\usepackage{amsmath}
\usepackage{amssymb}

\usepackage[
  backend=biber,
  style=numeric,
  sorting=none
]{biblatex}

\IEEEoverridecommandlockouts                              

\title{\LARGE \bf
BEVMAPMATCH: Multimodal BEV Neural Map
Matching for Robust Re-Localization of Autonomous Vehicles}

\author{Shounak Sural$^{*}$ and Ragunathan Rajkumar$^{*}$
\thanks{*Carnegie Mellon University, USA}}

\begin{document}

\maketitle
\thispagestyle{empty}
\pagestyle{empty}

\begin{abstract}
Localization in GNSS-denied and GNSS-degraded environments is a challenge for the safe widespread deployment of autonomous vehicles. Such GNSS-challenged environments require alternative methods for robust localization. In this work, we propose \setting{}, a framework for robust vehicle re-localization on a known map without the need for GNSS priors. \setting{} uses a context-aware lidar+camera fusion method to generate multimodal Bird's Eye View (BEV) segmentations around the ego vehicle in both good and adverse weather conditions. Leveraging a search mechanism based on cross-attention, the generated BEV segmentation maps are then used for the retrieval of candidate map patches for map-matching purposes.  Finally, \setting{} uses the top retrieved candidate for finer alignment against the generated BEV segmentation, achieving accurate global localization without the need for GNSS. Multiple frames of generated BEV segmentation further improve localization accuracy. Extensive evaluations show that \setting{} outperforms existing methods for re-localization in GNSS-denied and adverse environments, with a Recall@1m of $39.8\%$, being nearly twice as much as the best performing re-localization baseline. Our code and data will be made available at \url{https://github.com/ssuralcmu/BEVMapMatch.git}.

\end{abstract}
\vspace{-3mm}
\section{Introduction}
Accurate and reliable localization is a fundamental necessity in autonomous driving. Knowledge of the location of an ego vehicle within a map (referred to as re-localization) is fundamental for various Autonomous Vehicle (AV) tasks including global route planning, rule-abiding behavior planning and safe local path planning. Global Navigation Satellite System (GNSS) is one of the primary technologies used for localization in autonomous driving. While GNSS is widely adopted in autonomous driving systems for global positioning, its reliability degrades significantly in various challenging real-world environments, including urban canyons, tunnels, dense forests, mountain roads and adverse weather conditions. These regions are quite commonly seen during AV operations and motivate the need for developing methods that provide robust re-localization even in such challenging conditions. 

\begin{figure}
    \centering
    \includegraphics[width=\linewidth]{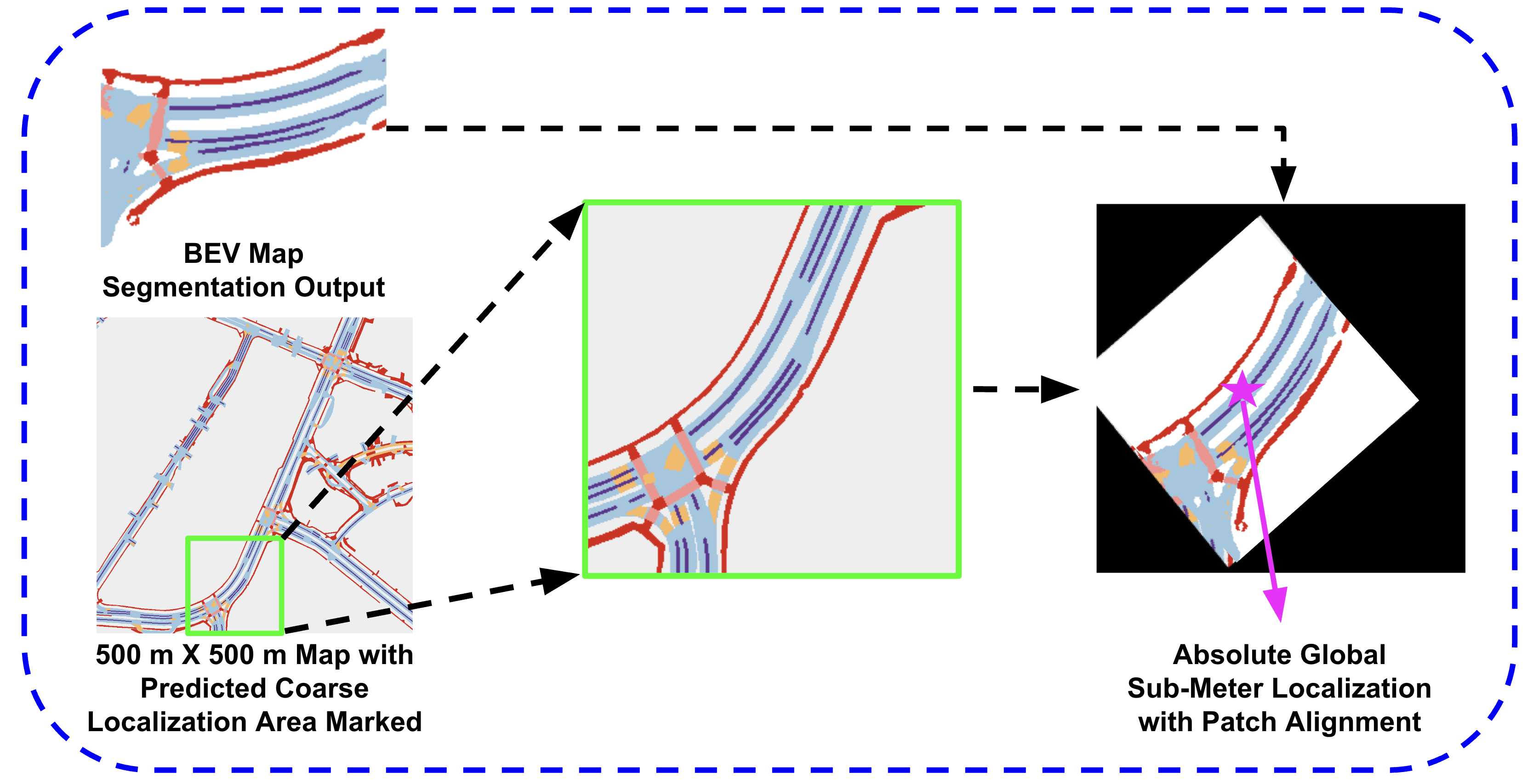}
     \caption{Sub-meter absolute re-localization with \setting{}.
Here, BEV segmentation maps are generated, followed by coarse matching that finds the most likely region, marked with a green square. This likely region is used to identify point-to-point correspondences against the segmented map. Finally, fine-grained pixel-level patch alignment yields accurate 3-DOF pose estimation (shown in pink).}
\label{fig:localization_example}
    \vspace{-7mm}
\end{figure}

While AVs can typically deal with a few meters of GNSS error and aim to correct such inaccuracies, GNSS errors in regions such as urban canyons can practically be as high as 177 m \cite{wen20213dlidaraidedgnss} or up to 200 m or higher in some operating contexts such as urban canyons \cite{wang2013sequential}. This typically happens because of a low count of visible GNSS satellites and multipath reflections.

A good amount of previous work addresses GNSS-denied localization with the use of dense HD lidar maps of the environment and using onboard lidar sensors to match and register generated point clouds against dense lidar maps for accurate localization. While localization using HD maps is effective, generating and maintaining 3D point cloud-based HD maps is expensive and difficult to scale. Furthermore, matching against dense lidar point clouds becomes computationally expensive when the pose uncertainty is large to begin with. Finally, point-to-point matching suffers from a degradation in performance in inclement weather conditions.

Bird's Eye View (BEV) is a well-established representation in autonomous driving that has various benefits for perception and planning tasks on AVs \cite{liu2022bevfusion, hu2023planningorientedautonomousdriving}. While lidars are inherently compatible with BEV, depth estimation-based feature extraction methods have also been extensively studied for single and multi-camera rigs. Recent camera-based methods show great promise \cite{zou2024m}. Furthermore, BEV representations can be aligned with maps containing both the structural and semantic information of lanes and road boundaries. This makes the BEV of the environment around an AV an attractive choice for use in visual localization.

In practice, localization that exploits the BEV representation has its own challenges. With many methods in the BEV space that use only cameras, their performance tends to drop in poor lighting conditions. Even with lidar+camera based methods, performance can degrade unless appropriate context is provided to aid the models in generalization to night-time scenes \cite{sural2024contextualfusion}. Nevertheless, even with good-quality BEV feature generation, it can be difficult to find a good match against a large global map. Given the large size of the search space, this matching process becomes computationally expensive and suffers from ambiguities.

In this work, we propose a GNSS-free localization framework called \setting{} that explicitly tackles these challenges. Figure \ref{fig:localization_example} describes an overview of our approach. \setting{} makes the following contributions:
\begin{itemize}
    \item A context-aware sensor fusion framework for BEV map segmentation that is robust to adverse operating conditions like inclement weather and low lighting.
    \item A cross-attention-based retrieval method that, without any GNSS priors, finds appropriate map patches for map-matching purposes.
    \item A fine-grained matcher that uses a neighborhood around a retrieved patch to find the best alignment within this region of the generated BEV map through feature-matching and subsequent homography estimation.
    \item Scalability to large maps with no GNSS information, while maintaining high precision by breaking down global localization into the two sub-problems of coarse patch retrieval and fine-grained alignment. 
    \item Highly accurate performance across various real-world autonomous driving scenarios that shows its effectiveness for GNSS-free global localization.
\end{itemize}

The rest of the paper is organized as follows. Section \ref{sec:related_work} discusses prior literature in the field of localization. Section \ref{sec:method} describes the \setting{} architecture and illustrates how the model is trained. Section \ref{sec:experiments} presents a detailed evaluation of \setting{}. Finally, we offer concluding remarks in Section \ref{sec:conclusion}.

\section{Related Work}\label{sec:related_work}
Accurate localization in GNSS-restricted environments has been studied extensively using multiple approaches. Prior work includes fusion of GNSS data with other onboard sensors such as Inertial Measurement Units (IMUs) and wheel odometry, point cloud registration against pre-collected lidar HD maps, aerial imagery-based localization and vision-based localization against map priors. These approaches differ primarily in terms of the sensors they rely on, the type of map they require, and the assumptions they make about the quality of GNSS availability which determine initial pose accuracy. These assumptions typically limit their robustness against adversities that AVs often face in the real world. 

\textbf{GNSS-based Fusion:} Many traditional localization methods rely on GNSS as the primary source for obtaining global position, followed by fusing with inertial measurements and wheel odometry using probabilistic filtering methods or factor graph optimization \cite{kaess_etal}. Kalman filters \cite{farag2022self}, particle filters \cite{8675596} and Monte Carlo Localization \cite{8910578} are also commonly used to combine these measurements, achieving high accuracy when GNSS measurements are reliable \cite{9833300}. A fusion of vision-based methods, GNSS and Kalman-filter-based Inertial Navigation System (INS) has been used to effectively perform reliable localization over long distances \cite{extended_vins_mono}. However, the fundamental assumption in several of these methods is the availability of continuous GNSS and its localization error within a few meters. In commonly seen environments such as tunnels, urban canyons or underpasses, GNSS measurements can be unreliable, leading to large localization errors that cannot be corrected only by using these sensors.

\textbf{Lidar HD Map-based Localization:} To overcome the limitations of GNSS-based methods, many studies have explored map-based localization, where onboard sensor data is registered against a pre-built map. Lidar-based re-localization techniques that use pre-built 3D HD maps consisting of point clouds can be very accurate \cite{9833300}. Traditional scan matching methods such as Iterative Closest Point (ICP) that align the onboard lidar generated point cloud to the HD map is commonly used \cite{7995900}. ML models such as SegMatch \cite{7989618} have also been used for improved accuracy for the same task. However, despite their effectiveness, such dense lidar HD maps are expensive to develop, maintain and scale.

\begin{figure*}
    \centering
    \includegraphics[width=0.9\linewidth]{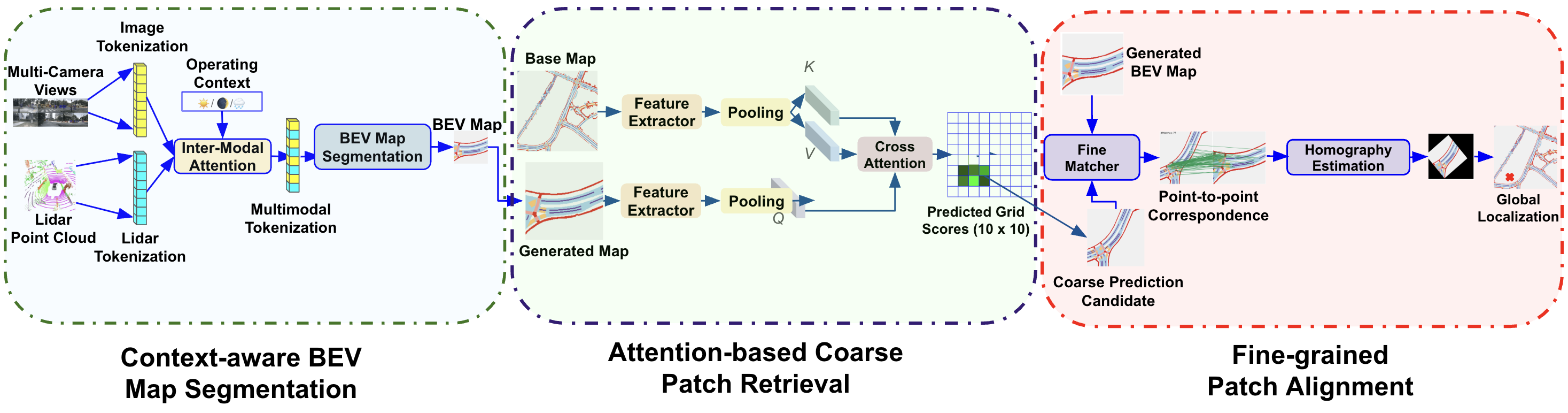}
    \vspace{-3mm}
    \caption{Overview of the BEVMapMatch Architecture}
    \label{fig:bevmapmatch_arch}
    \vspace{-5mm}
\end{figure*}
Another line of work uses satellite imagery instead of 3D lidar maps for matching. In this case, camera images are fed into deep learning networks that learn cross-view representations, typically in the visual BEV space for matching \cite{7989239}. However, differences between ground and aerial views, possible obstruction by structures, vegetation and other vehicles as well as temporal changes with new buildings being built over time can lead to mismatches. 

\textbf{Camera-based Localization:} Vision-based localization methods that use cameras for matching against existing maps are attractive as a low-cost localization option. Nevertheless, most existing methods such as MapLocNet \cite{10802757} that deal with localization in GNSS-degraded regions assume a GNSS prior that effectively limits matching and errors to a small spatial neighborhood. This refines coarse GNSS estimates but does not have the ability to reliably recover a global position from scratch required for re-localization.

Bird's Eye View (BEV) representation learning from multiple camera images and potentially lidar data have proven to be highly effective in various tasks for object detection, drivable region segmentation and map understanding \cite{liu2022bevfusion, wang2023unitrunifiedefficientmultimodal}. They have also been effective for downstream tasks such as planning with the use of end-to-end methods \cite{hu2023planningorientedautonomousdriving}. However, even with multiple cameras, and even lidars, their performance can degrade at night \cite{liu2022bevfusion}. \setting{} deals with this challenge using context-aware fusion during BEV map segmentation generation. Methods such as OrienterNet \cite{sarlin2023orienternet} perform camera-only localization, but they rely on reliable information about building shapes to obtain the match. With monocular camera images, their reliability goes down at night in practice, since datasets like KITTI used for evaluation have camera images only in good weather and lighting conditions. 

\textbf{Challenges to be Addressed:} Unlike methods such as MapLocNet \cite{10802757} that deal with a small search space of a few meters to sometimes $30m$ for GNSS accuracy, \setting{} explicitly targets global localization over a much larger map without any pose prior. This is referred to as the re-localization problem, also called the kidnapped robot problem in the literature. While more challenging, re-localization has the primary advantage of being a much more generalizable method that works without the need for GNSS, exactly what is required in challenging scenarios for AVs such as urban canyons. For safe AV operations, no matter how large the map is, localization still has to be accurate at a lane or sub-lane level. Searching for an exact pixel in such a large search space of a map, which is say $500m\times500m$ can be computationally intractable. Due to the fact that GNSS is highly unreliable in challenging environments, the heading of the vehicle with respect to the map is also unknown. When searching for an appropriate match in the larger map, the lack of heading information becomes an additional degree of freedom that increases the search space potentially by multiple orders of magnitude.
An alternative to address the map-matching problem is to use deep-learning feature-based matching techniques for finding point-to-point correspondences in images. Traditional methods (such as SIFT) \cite{lowe2004distinctive}, rotation-invariant methods (like ORB \cite{6126544}) and transformer-based methods such as ROMA \cite{edstedt2023romarobustdensefeature} and LOFTR \cite{sun2021loftrdetectorfreelocalfeature} can be utilized. While these approaches work for many types of images with multiple views, with regards to our goal, the map image is simply too large, and directly using such methods fails with severe mismatches. 

In summary, while many map-matching methods exist, their robustness in challenging GNSS-denied environments is still lacking. \setting{} aims to fill this gap by introducing the notion of context-aware fusion in cross-modal attention architectures for robustness in adverse conditions. Subsequently, it tackles the problem of precise localization in a large map using a hierarchical approach. The first stage develops a cross-attention-based matcher and treats it as a retrieval task aiming to predict the most likely candidates. Within the best matches, using image-to-image correspondences, it finds a precise match, obtaining sub-meter-level accuracy in many cases.
\vspace{-1mm}
\section{The \setting{} Architecture}\label{sec:method}
In this section, we present the architecture of \setting{} and describe how \setting{} achieves accurate localization without the need for GNSS priors in AVs. First, we present a context-aware BEV representation learning framework, followed by an attention-based patch retrieval method for re-localization up to a level of coarse map patches. Finally, we discuss a method that uses the top retrieved candidate for precise patch alignment, that enables global localization.
\vspace{-2mm}
\subsection{Problem Definition}
\setting{} addresses the problem of re-localization in autonomous driving, with localization under adverse conditions on a given map where standard methods of GNSS-aided localization have failed. Specifically, we aim to solve the following optimization problem: 

Let $\mathbf{z}_{1:T}=\{\mathbf{I}^{1:N_c}_{1:T},\,\mathbf{P}_{1:T}\}$ denote the sensor data history over timesteps
$t=1,\dots,T$, where $\mathbf{I}^{j}_t$ is the $j^{th}$ camera image for $j=1,\dots,N_c$ and $\mathbf{P}_t$ is the LiDAR point cloud at time $t$. We define $\mathcal{M}$ as the known geo-referenced map, and $\mathbf{x}_T=(x_T,y_T,\theta_T)\in SE(2)$ (Special Euclidean group) is the ego vehicle's 3-DoF pose at time T.
\setting{} estimates the pose $\mathbf{x}_T$ by optimizing
\begin{equation}
\hat{\mathbf{x}}_T=\arg\max_{\mathbf{x}_T} p(\mathbf{x}_T \mid \mathbf{z}_{1:T}, \mathcal{M})    
\end{equation}
As shown in Figure \ref{fig:bevmapmatch_arch}, \setting{} performs the above optimization in three stages: (1) BEV map segmentation, (2) coarse patch retrieval and (3) fine patch alignment.

\subsection{Context-Aware BEV Representation Learning}
Multi-view camera images covering all regions around an AV and a lidar point cloud with similar coverage are provided as inputs to \setting{}. Context information, specifically whether it is day or night-time and whether it is raining, is also provided as an additional input to the network, to help in stabilizing performance across various operating conditions \cite{sural2024contextualfusion,sural2024contextvlm}. We refer to our BEV representation learning and map segmentation model as Context-Aware Fusion (CAF). CAF builds upon the well-known UniTR model \cite{wang2023unitrunifiedefficientmultimodal} to incorporate environmental context into multi-sensor fusion for downstream tasks. 

CAF uses a Vision Transformer-based tokenizer for processing camera images and the MVF tokenizer \cite{zhou2020end} for lidar points. Before the inter-modal attention stage in Figure \ref{fig:bevmapmatch_arch}, learnable context-based fusion layer adaptively learns to prioritize camera and lidar features explicitly depending on how effective each sensor type is at night and in rain. Lidar, for example, being an active sensor, is equally effective during day and night. However, camera images tend to capture fewer details at night resulting in a drop in performance \cite{liu2022bevfusion,sural2024contextualfusion, sural2024contextvlm}. Lidars also typically generate noisier point clouds in rain and an appropriate relative weighting is learnt with our context-aware fusion module. The remainder of the inter-modal attention stage follows that of UniTR \cite{wang2023unitrunifiedefficientmultimodal}. Once the representations are learnt, a BEV map segmentation head consisting of convolutional layers, followed by additional 1x1 convolutions to finally produce per-class logits for each pixel. At this point, a map segmentation of the region around the ego vehicle is obtained. Focal loss is then used to train the model for pixel-wise semantic segmentation of a 100m$\times$100m region around the ego vehicle with six classes of interest: Drivable Area, Pedestrian Crossing, Walkway, Stop Line, Carpark Area and Divider. 

\subsection{Coarse Map Localization via Patch Retrieval}
Once the map segmentation has been trained well, the next step is to use the segmentations for map matching-based re-localization. We start with a base map of the region within which the ego vehicle can be located anywhere. The base map consists of road structure information such as lane information, intersections, crosswalks and stop lines. This map has been converted offline into a format which visually looks similar to the map segmentation image to minimize style transfer gap during matching. The generated BEV segmentation occupies a much smaller region compared to the base map, which is taken as $500m\times500m$ in most of our experiments. We choose this size for the geo-referenced map since $500m$ is substantially larger than GNSS-degraded priors used in prior work (which are tens of meters). While we choose a size of $500m\times500m$ for controlled evaluation, our proposed \setting{} method is not tied to this map size and can be scaled to much larger maps by using the same hierarchical strategy.

Both the segmented image and the base map image are passed through a shared DINOv2-based backbone  \cite{oquab2024dinov2learningrobustvisual} to serve as a general feature extractor. The backbone is frozen and outputs patch token embeddings which are converted into a spatial feature map. For the base map, features are pooled to generate a $10\times10$ grid. To create this \textit{base grid}, a 3x3 neighborhood is chosen for each grid cell and averaged to capture the local context for each grid cell. The generated BEV segmentation map is passed through the same DINOv2 network and the output \textit{segmented map tokens} are used for matching against the base grid. We leverage a cross-attention mechanism which effectively treats the generated BEV mask as the query and uses the base grid as corresponding keys and values. Furthermore, a learnable 2D positional embedding is appended to both the BEV and base grid tokens. Multi-headed attention with 8 heads is used to learn a query-conditioned representation. Finally, a fully connected linear layer is used to output 100 logits, one for each base grid cell, representing the coarse localization scores. This procedure is formulated as a retrieval task with best matched positions being returned along with corresponding scores.

The coarse map localization stage primarily uses a binary cross entropy loss for training the network. However, such a loss does not take distance into account. Hence, a prediction for a cell which is the neighbor of the top-most matching cell is considered a negative and not a strong enough signal. To deal with this issue, we include an additional term to our loss function as shown in (2), which uses a distance-aware loss with a Gaussian-based soft distribution around the target cell. From our experimentation, this term, however, is not strong enough by itself to train the neighbor in a stable manner. The sum of these two losses is therefore used for training the coarse localization stage as described below. 

For each sample $b$ in a mini-batch of size $B$, we predict logits $\mathbf{z}_b\in\mathbb{R}^{N}$ over a $G\times G$ base grid, where $N=G^2$.
Let $\mathbf{y}_b\in\{0,1\}^{N}$ be the one-hot target grid with target center cell $k_b$ and 2D coordinate $\mathbf{c}_b$.
We optimize
\begin{equation}
\mathcal{L}
=\mathcal{L}_{\text{BCE}}+\lambda\,\mathcal{L}_{\text{Dist}}
\end{equation}
\vspace{-3mm}
where,
\vspace{-2mm}
\begin{equation}
\mathcal{L}_{\text{BCE}}
=\frac{1}{BN}\sum_{b=1}^{B}\sum_{i=1}^{N}\mathrm{BCEWithLogits}(z_{b,i},y_{b,i}),
\end{equation}
and the distance-based term $L_{Dist}$ given in (4) rewards proximity to the ground truth cell even if there is no exact match.
\begin{equation}
\mathcal{L}_{\text{Dist}}
=-\frac{1}{B}\sum_{b=1}^{B}\sum_{i=1}^{N}\tilde{y}_{b,i}\log (\mathrm{softmax}(\mathbf{z}_{b,i}))
\end{equation}
The soft target $\tilde{\mathbf{y}}_b$ is a normalized Gaussian over base grid cells $\mathbf{p}_i\in\mathbb{R}^2$, centered at the ground truth cell center $\mathbf{c}_b$:
\begin{equation}
\tilde{y}_{b,i}
=\frac{\exp\!\left(-\|\mathbf{p}_i-\mathbf{c}_b\|_2^2/(2\sigma_d^2)\right)}
{\sum_{j=1}^{N}\exp\!\left(-\|\mathbf{p}_j-\mathbf{c}_b\|_2^2/(2\sigma_d^2)\right)} .
\end{equation}
\subsection{Fine-Grained Patch Alignment}
From the coarse matching stage, we obtain the top predicted $1\times1$ cell within the $10\times10$ base grid. An analysis of the error distribution from the coarse matcher (more details in Section \ref{sec:experiments}) reveals that, in most cases, the predicted cell either matches the exact cell or the prediction is one cell away from the ground truth. This is understandable since the match is being searched on a $100m\times100m$ map, while the predicted cell is $50m\times50m$. Hence, a neighboring cell can also be positive. To deal with this, we take the $3\times3$ neighborhood around the top predicted cell, which is $150m\times150m$ and aim to find an appropriate patch alignment of the generated BEV segmentation map within this region. Note that the predictions of the best matching $2\times2$ or $3\times3$ cells are also valid alternatives towards solving this problem. However, based on our experience, the stronger requirement of predicting the top $1\times1$ cell works much better.

While coarse localization in a much larger region is a harder problem to solve, fine-grained alignment is a problem that has been relatively well studied. Especially, image processing methods that use traditional feature extractors to find a homographic transformation between two images is quite common. Along these lines, we adopt the efficient LoFTR method \cite{wang2024efficient} for our task. A pre-trained version of this model, which is used for map matching across a wide variety of tasks, has been introduced in MatchAnything \cite{he2025matchanything}. 

The two input images are processed with a CNN backbone for multi-scale feature extraction. A transformer-based network is used to transform these features. An efficient token aggregation method is used to reduce the computational load of full transformer attention across all tokens. This step is followed by a correlation method that produces a similarity matrix, which then selects correspondences between the two images. These correspondences are then refined to find precise matches at sub-pixel precision. Finally, with the obtained matches, RANSAC is used to estimate a homographic transformation between the two images. This transformation is used to find the projection of the center of the generated BEV segmentation map (i.e., location of ego vehicle) on the $3\times3$ predicted map patch. Once the exact pixel is obtained in this representation, the corresponding location in the much larger area map can be obtained, accurate to a pixel level. The final outcome is an absolute global localization of the AV in a large map without the use of GNSS priors.
\begin{table}[t]
\centering
\caption{Coarse localization accuracy comparison on the NuScenes-based dataset (1 frame).}
\vspace{-3mm}
\label{tab:coarse_loc_acc}
\scriptsize
\setlength{\tabcolsep}{4pt}
\begin{tabular}{c|c|ccc}
\hline
\textbf{Model} & \textbf{Segmentation} & \textbf{Seg. IOU} & \textbf{Top 1$\times$1} & \textbf{Top 3$\times$3} \\
\hline
BEVMapMatch & BEVFusion~\cite{liu2022bevfusion} & 62.7 & 38.6 & 81.1 \\
BEVMapMatch & CAF (Ours)             & \textbf{75.0} & 46.7 & 86.4 \\
BEVMapMatch & Ground Truth                     & 100.0 & 50.9 & 89.9 \\
\hline
\end{tabular}
\end{table}
\vspace{-5mm}
\begin{table}[]
\centering
\caption{Effect of temporal fusion on coarse localization performance of \setting{}.}
\vspace{-3mm}
\label{tab:coarse_temporal_fusion}
\begin{tabular}{ccc}
\hline
\textbf{Number of frames} & \textbf{Top 1$\times$1 Acc. (\%)} & \textbf{Top 3$\times$3 Acc. (\%)} \\
\hline
1  & 46.68 & 86.37 \\
2 & 50.76 & 88.77 \\
4 & 53.19 & \textbf{88.90} \\
8 & \textbf{53.46} & 88.65 \\
\hline
\end{tabular}
\vspace{-3mm}
\end{table}
\begin{table}[t]
\centering
\caption{\setting{} backbone ablation with fixed basemap input resolution of $224 \times 224$.}
\vspace{-3mm}
\label{tab:bevmapmatch_backbone_ablation}
\begin{tabular}{lcc}
\hline
\textbf{Backbone Used} & \textbf{Top 1$\times$1 Acc. (\%)} & \textbf{Top 3$\times$3 Acc. (\%)} \\
\hline
ResNet-34 \cite{he2015deepresiduallearningimage}  & 17.68 & 63.18 \\
DINO-v2-Base \cite{oquab2024dinov2learningrobustvisual}  & 41.15 & 86.12 \\
DINO-v2-Large \cite{oquab2024dinov2learningrobustvisual} & \textbf{46.68} & \textbf{86.37} \\
\hline
\end{tabular}
\vspace{-4mm}
\end{table}

\begin{table*}[t]
\centering
\caption{Localization performance analysis with absolute recall for precise lane-level localization on NuScenes (val). Scaled recall metrics are used for a fair comparison against existing methods.}
\label{tab:combined_abs_scaled_recall}
\setlength{\tabcolsep}{4.5pt}
\renewcommand{\arraystretch}{1.12}
\begin{tabular}{c |c | cc || cccc}
\hline
\textbf{\multirow{2}{*}{Method}} & \textbf{\multirow{2}{*}{Venue}}
& \multicolumn{2}{c||}{\textbf{Absolute Recall (\%)}} 
& \multicolumn{4}{c}{\textbf{Scaled Recall (\%)}} \\
\cline{3-8}
& & \textbf{@1m} $\uparrow$ & \textbf{@2m} $\uparrow$
 & \textbf{Eq.@1m} $\uparrow$ & \textbf{Eq.@2m} $\uparrow$ & \textbf{Eq.@5m} $\uparrow$ & \textbf{Eq.@10m} $\uparrow$ \\
\hline

\multicolumn{8}{c}{\textit{Prior localization methods}} \\
\hline
OrienterNet~\cite{sarlin2023orienternet} & CVPR 2023  & 5.83 & 18.92 & 5.83 & 18.92 & 52.83 & 66.21 \\
U-BEV~\cite{camiletto2024u}     & IROS 2024          & 16.89 & 41.60 & 16.89 & 41.60 & 71.33 & 83.46 \\
MapLocNet~\cite{10802757}   & IROS 2024    & 20.10 & 45.54 & \textbf{20.10} & \textbf{45.54} & \textbf{77.70} & \textbf{91.89} \\
\hline

\multicolumn{8}{c}{\textit{Ours}} \\
\hline
BEVMapMatch (1-frame) & - & 38.52 & 52.12 & 57.04 & 60.21 & 71.36 & 83.00 \\
BEVMapMatch (2-frame) & - & 39.33 & 53.57 & 58.43 & 61.71 & 73.43 & 84.99 \\
BEVMapMatch (4-frame) & - & \textbf{39.82} & \textbf{54.04} & \textbf{58.93} & \textbf{62.40} & 74.51 & 85.16 \\
BEVMapMatch (8-frame) & - & 39.55 & 53.97 & 58.84 & 62.21 & 74.35 & 85.10 \\
\hline
\end{tabular}
\vspace{-5mm}
\end{table*}
\vspace{3mm}
\section{Evaluation} \label{sec:experiments}
\subsection{Dataset and Experimental Settings}
We adopt the widely used NuScenes dataset \cite{caesar2020nuscenes} to keep our results reproducible. However, NuScenes by default does not come with a framework that supports localization. However, for each sample, it does contain location information, which can be identified against a standardized base map provided. For tasks such as BEV segmentation, a map region around the ego vehicle can be used for training. We use a map size of $500m\times500m$ for our experiments as motivated earlier in Section \ref{sec:method}. Within this map, we perturb the 2-DOF location of the vehicle by a random amount in the range $[-200m,+200m]$ to both X and Y coordinates, resulting in a maximum absolute GNSS error of $282.8 m$ within the bounds of these maps to set up the task of GNSS-free localization. Hence, the location of the vehicle is distributed within the $500m\times500m$ map. The orientation of the ego vehicle does not align with the map boundaries by default and hence can be in any random direction. We use this observation to create our dataset to test GNSS-free localization performance. Our dataset uses the exact distribution of NuScenes with 28,130 samples for training and 6,019 for validation.

\subsection{Segmentation-guided patch retrieval performance}
The first stage of \setting{} performs BEV map segmentation for a $100m\times100m$ region around the ego vehicle. This is followed by the second stage, a coarse retrieval pipeline. We train our model to accurately predict the most likely cell out of the $10\times10$ base grid. Table \ref{tab:coarse_loc_acc} shows the performance of the map matcher in \setting{}. We first show the IOU of the first segmentation stage, followed by the accuracy of predicting the exact cell and the accuracy of the top $3\times3$ region being an exact match. $3\times3$ accuracy is relevant because we are trying to find a match for a $100m\times100m$ map against a hundred $50m\times50m$ cells in the base grid. Hence, a neighboring cell in most cases is a valid match, and significantly aids the next fine-grained matching stage. For these metrics, we perform a comparison of segmentations generated by a state-of-the-art BEVFusion model and our Context-Aware Fusion (CAF) model. We see that for identifying the correct top $3\times3$ region, our method obtains an accuracy of $86.4\%$, which is very close to the performance of our model on using ground truth segmentations. Our CAF model shows significant improvement over state-of-the-art map segmentation baselines such as BEVFusion.

\begin{figure}[]
    \centering
    \includegraphics[width=0.9\linewidth]{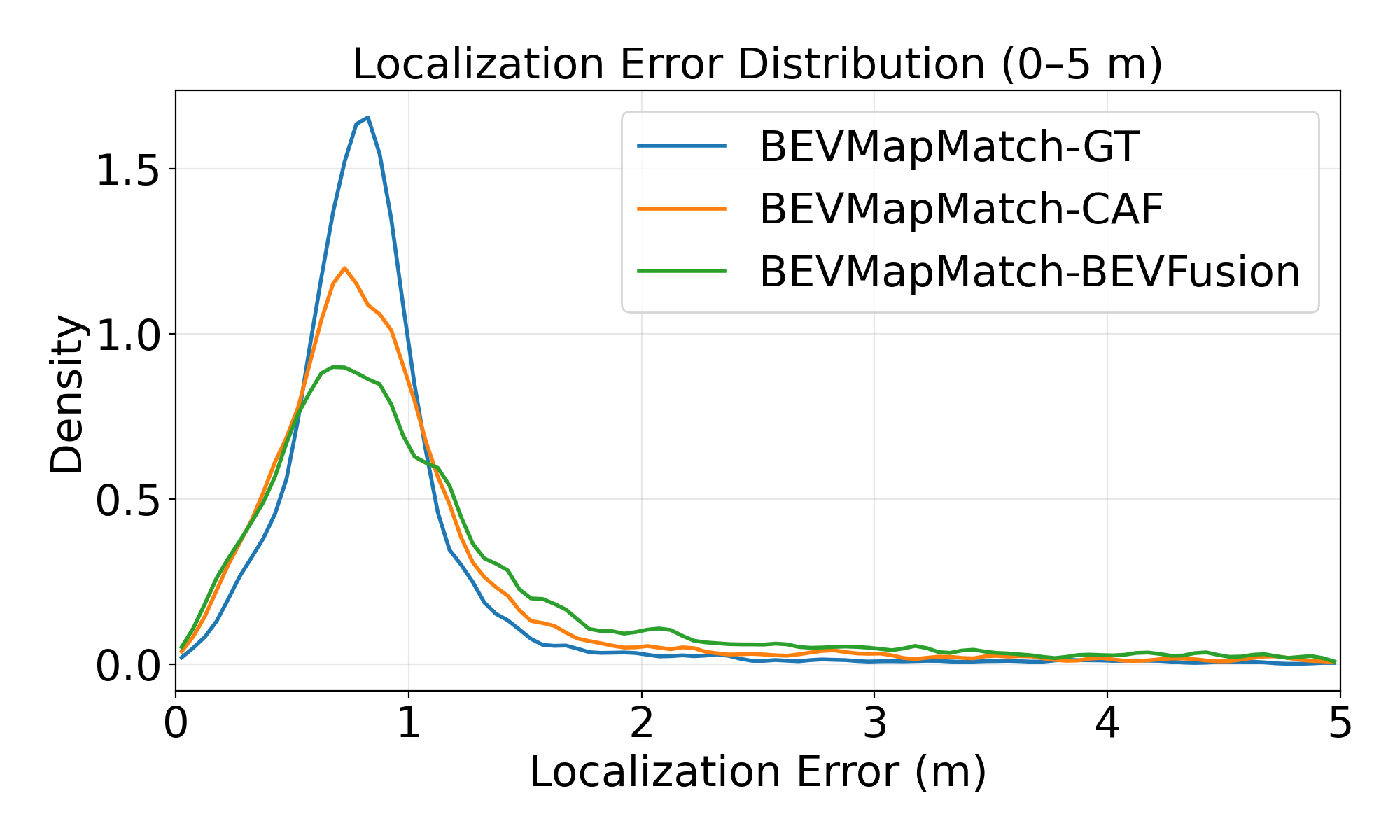}
    \vspace{-4mm}
    \caption{Comparison of \setting{} performance across choice of segmentation baselines and the ground truth.}
    \label{fig:fine_error_dist_plot}
        \vspace{-4mm}
\end{figure}
The patch retrieval performance improves significantly on using more frames over time. Table \ref{tab:coarse_temporal_fusion} shows the effect of temporal fusion on coarse localization performance with the best performing segmentation model. Accumulating history to create richer and better matching representations significantly improves coarse localization accuracy. Using 2 frames boosts the relative performance by 8.7\% over using only 1 frame, going up to 14.5\% with 8 frames. Table \ref{tab:coarse_temporal_fusion} also shows that the top $1\times1$ accuracy increases significantly up to a total of 4 frames and then relatively plateaus. The top $3\times3$ accuracy plateaus at 2 frames. However, depending on operational platform constraints, using either 2 or 4 frames can be hugely beneficial in improving coarse localization performance. With 4 segmentation frames, the top $1\times1$ accuracy of 53.19 exceeds even that of using the ground truth segmentation with 1 frame and nearly matches the top $3\times3$ accuracy of using the ground truth. In the annotated version of NuScenes \cite{caesar2020nuscenes}, every frame is recorded 0.5 seconds apart. The motion of the ego vehicle across frames allows more context for better map matching. 

In our experiments, we use the DINO-v2 (Large) model \cite{oquab2024dinov2learningrobustvisual} as our feature extraction backbone for the coarse localization module starting from the BEV map segmentations as input. Our choice of such a backbone is validated in Table \ref{tab:bevmapmatch_backbone_ablation} with an ablation study against other candidates. We see that the choice of frozen DINO-v2 as the backbone provides a major boost over using Resnet-based feature extractors. This improvement primarily happens because ResNet is trained on ImageNet \cite{he2015deepresiduallearningimage}, while DINOv2 is trained with self-supervision on a huge 142 million dataset, namely LVD-142M \cite{oquab2024dinov2learningrobustvisual}, and is geared towards generating general-purpose visual features. 

\begin{figure}[]
    \centering
    \includegraphics[width=0.9\linewidth]{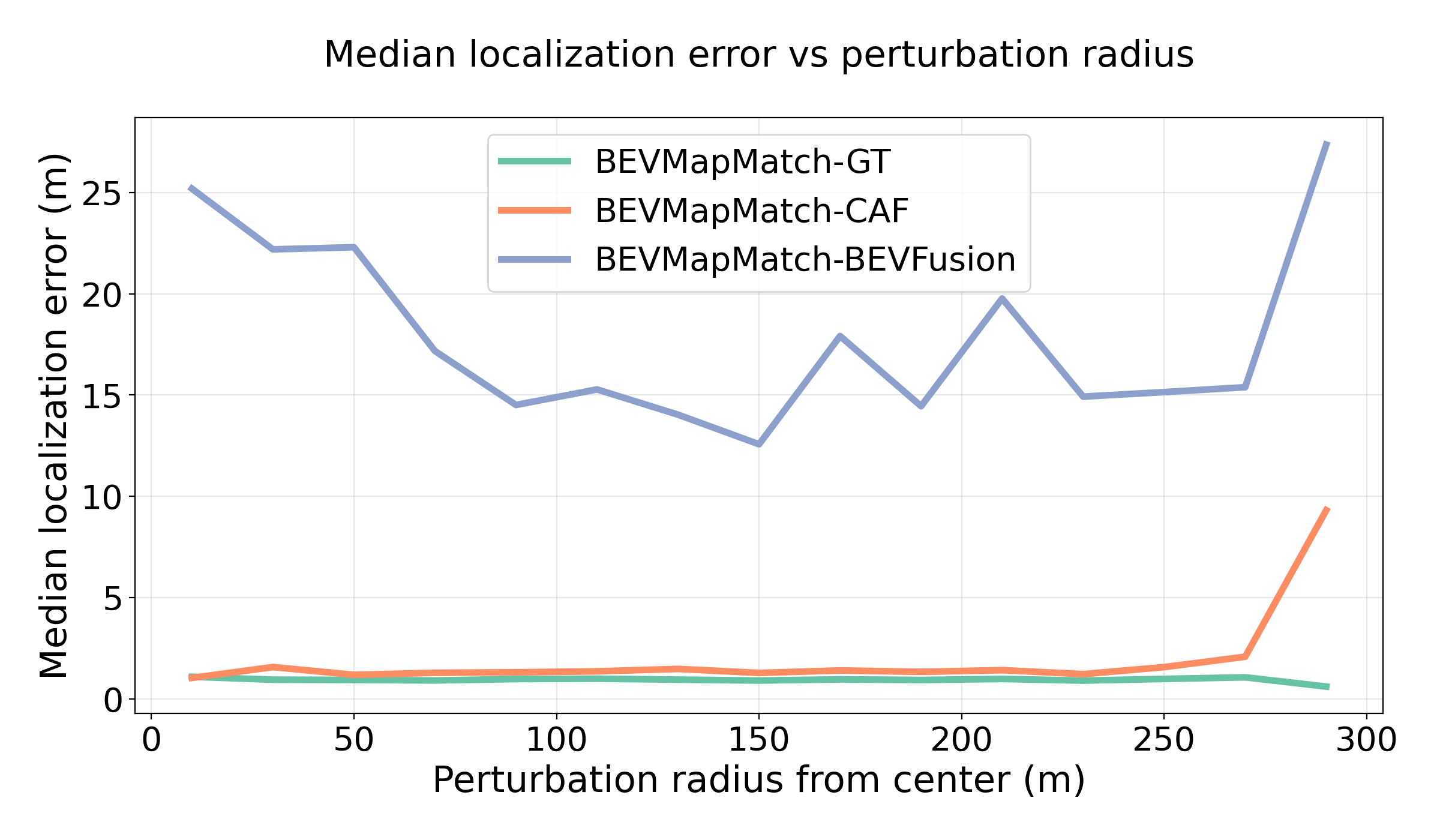}
    \vspace{-5mm}
     \caption{Robustness of BEVMapMatch with random GNSS perturbation. \setting{}-CAF stays stable over varying GNSS errors since it is retrieval-based and does not use GNSS priors.}
    \label{fig:perturb_robustness_plot}
        \vspace{-5mm}
\end{figure}
\begin{figure*}[t]
    \centering
    \includegraphics[width=0.95\linewidth]{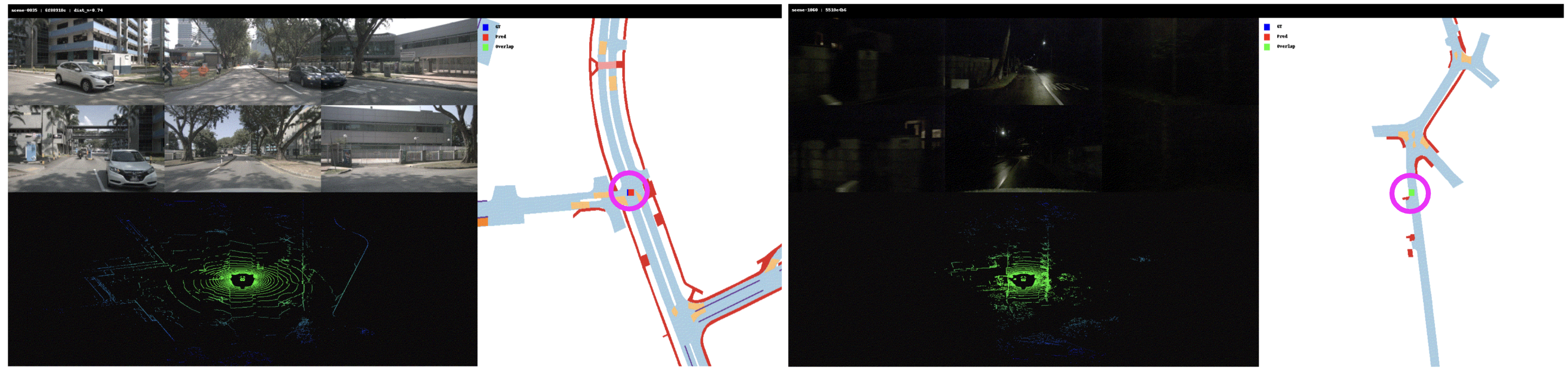}
    \vspace{-3mm}
    \caption{Examples of localization at day and night-time with \setting{}. Camera and lidar views shown (left) along with its localization output on a NuScenes base map (right). Within pink circles, red squares mark the predictions, blues mark ground truth and greens mark full overlap.}
    \vspace{-2mm}
    \label{fig:localization_examples_viz}
\end{figure*}
\vspace{-2mm}
\subsection{Post-Retrieval Fine-grained Patch Alignment}
Using the outputs of the coarse localization stage, we form a $3\times3$ cell region from the $10\times10$ cell region map.  With this, we perform fine-grained localization within the base map, aiming to predict the exact pixel that marks the location of the ego vehicle. Here, our goal is to predict an exact pixel from a $1000\times1000$ base map image, with each pixel spanning $0.5m\times0.5m$. Table \ref{tab:combined_abs_scaled_recall} shows the performance of \setting{}, compared against prior methods such as OrienterNet \cite{sarlin2023orienternet}, U-BEV \cite{camiletto2024u} and MapLocNet \cite{10802757} for re-localization. Unlike these methods which only use camera images, \setting{} also incorporates lidar sensors. Table \ref{tab:combined_abs_scaled_recall} evaluates two metrics: absolute and scaled recall. 

For absolute recall, the values for prior methods are taken from MapLocNet \cite{10802757} since this is the closest paper to ours which also uses a similar NuScenes-derived dataset. Absolute recall @\textit{d} m shows the fraction of samples where localization is accurate to within \textit{d} meters. Scaled recall is similar but accounts for the difference in our dataset and the one used in MapLocNet \cite{10802757}, both derived from NuScenes. The MapLocNet version of GNSS-perturbed NuScenes has a perturbation range of $[-30m,30m]$ while ours is $[-200m,200m]$, making our search space 6.67 times larger in each dimension. Hence, as an example, $2 m$ in their model corresponds to $13.34 m$ in ours. This relative difference is accounted for when calculating the scaled recall metrics where the thresholds are scaled equivalently. Note that the MapLocNet version has a smaller overall map size of $128 m$ versus our $500 m$ and has a much smaller variation in GNSS heading error ($[-30 ^\circ,+30 ^\circ]$), while ours can have any orientation change ($[-180 ^\circ,+180^\circ]$). This additional complexity actually makes our problem much harder, but for a simple and fair comparison, we do not account for such variations in this table. 

\begin{table*}[t]
\centering
\caption{Localization performance (\%) of \setting{} across weather/lighting subsets of NuScenes.
\textbf{1$\times$1 accuracy} = Exact 1x1 GT Cell Matched, \textbf{3$\times$3 accuracy} = Exact 3x3 GT Cell Matched.
R@X m is recall at X meters of absolute localization error}
\vspace{-2mm}
\label{tab:context}
\scriptsize
\setlength{\tabcolsep}{3pt}
\begin{adjustbox}{max width=\textwidth}
\begin{tabular}{cc|cccc|cccc|cccc|cccc}
\toprule
\multirow{2}{*}{\textbf{Model}} & \multirow{2}{*}{\textbf{\# Frames}}
& \multicolumn{4}{c|}{\textbf{Day}}
& \multicolumn{4}{c|}{\textbf{Night}}
& \multicolumn{4}{c|}{\textbf{Rain}}
& \multicolumn{4}{c}{\textbf{Overall}} \\
&
& \textbf{1$\times$1} & \textbf{3$\times$3} & \textbf{R@1m} & \textbf{R@2m}
& \textbf{1$\times$1} & \textbf{3$\times$3} & \textbf{R@1m} & \textbf{R@2m}
& \textbf{1$\times$1} & \textbf{3$\times$3} & \textbf{R@1m} & \textbf{R@2m}
& \textbf{1$\times$1} & \textbf{3$\times$3} & \textbf{R@1m} & \textbf{R@2m} \\
\midrule
BEVMapMatch-BEVFusion & 1
& 41.03 & 83.95 & 29.53 & 42.62
& 36.17 & 82.09 & 14.01 & 22.87 
& 28.95 & 68.33 & 17.20 & 28.76
& 38.59 & 81.13 & 25.91 & 38.25 \\
\midrule
\multirow{4}{*}{BEVMapMatch-CAF}
& 1
& 50.46 & 89.15 & 42.86 & 56.91
& 42.02 & 87.59 & 23.94 & 36.52 
& 32.42 & 74.25 &  28.88 & 41.67
& 46.68 & 86.37 & 38.52 & 52.12 \\
& 2
& 54.33 & 90.90 & 43.58 & 57.89
& 46.28 & 92.38 & 25.53 & 39.89
& 37.41 & \textbf{78.38} & 29.61 & 44.55
& 50.76 & 88.77 & 39.33 & 53.57 \\
& 4
& 56.45 & 91.17 & \textbf{43.91} & 56.09 
& \textbf{49.29} & \textbf{92.91} & \textbf{26.11} & \textbf{39.25}
& 40.32 & 77.91 & \textbf{30.92} & 45.39
& 53.19 & \textbf{88.90} & \textbf{39.82} & \textbf{54.04} \\

& 8
& \textbf{56.97} & \textbf{91.51} & 43.86 & \textbf{58.14}
& 47.87 & 92.20 & 22.87 & 37.94 
& \textbf{40.60} & 75.47 & 30.67 & \textbf{46.47} 
& \textbf{53.46} & 88.65 & 39.55 & 53.97  \\
\bottomrule
\end{tabular}
\end{adjustbox}
\vspace{-6mm}
\end{table*}

In terms of absolute metrics, \setting{} outperforms existing methods, being nearly twice as effective as the best performing baseline MapLocNet \cite{10802757} for absolute recall@1m. For the more lenient yet fairer scaled recall metrics, \setting{} also shows performance improvement, while having slightly worse distribution for recall at higher error thresholds. This is partly due to the additional complexity increase with our dataset that is not captured in the scaled recall metric. A further analysis of the relative density of our fine-grained localization error distribution across design choices of BEVMapMatch is illustrated in Figure \ref{fig:fine_error_dist_plot}. GT refers to the use of ground truth segmentations, which is only for demonstration of the peak attainable performance. CAF and BEVFusion are segmentation backbones that can be used in practice for deployment. Figure \ref{fig:fine_error_dist_plot} shows that \setting{} with our CAF segmentation model, followed by our matching models, results in a median error peak that is below 1 m considering all samples in the 0-5 m error range. Thus, in the most cases, \setting{} is able to localize accurately at sub-meter levels without the use of GNSS priors in a $500m\times500m$ map.

Figure \ref{fig:perturb_robustness_plot} shows the distribution of median localization error as GNSS perturbation variation is varied. While the BEVFusion-based segmentation model has a top $3\times3$ accuracy of $81.1\%$ as opposed to $86.4\%$ for our CAF model in Table \ref{tab:bevmapmatch_backbone_ablation}, a major difference manifests when performing finer-grained pixel-level localization with feature matching. That is, the quality of segmentation matters a lot for fine-grained localization. However, CAF performs very similar to the ground truth. Another key insight from Figure \ref{fig:perturb_robustness_plot} is that \setting{}-CAF typically has a stable median localization error no matter how off the GNSS is. The increase in error for the tail at the end is because of the GNSS going to the edge of the map, effectively resulting in a blank white area being used as the search query for retrieval in some cases causing ambiguous returns. This phenomenon happens because we do not rely on GNSS priors and design our method in a way that can handle such large abnormalities in GNSS, commonly observed in GNSS-denied areas.
\vspace{-2mm}
\subsection{Operating Context-based Analysis}
Table \ref{tab:context} shows an analysis of the segmentation method variations of \setting{} evaluated across varying operating conditions. While our models outperform existing baselines, all models face a relative performance drop at night and in rain. Rain can significantly affect lidar point clouds, which results in road edges being blurred. Similarly, night-time causes a drop in performance due to poor visibility of the road structure with cameras. The performance drop for fine-grained localization is therefore even more pronounced at night. Note that the generated BEV segmentations are used twice in \setting{}, both in the coarse- and fine-grained stages. In other words, the success of the fine-grained stage depends on the success of both the coarse- and fine-grained stages. A moderate relative drop in performance is observed for $3\times3$ coarse matching with adverse conditions, but a more pronounced drop occurs at the fine-grained localization stage. Figure \ref{fig:localization_examples_viz} shows sensor data in day and at night along with \setting{} localization performance. Note that the visualized map is $200 m \times 200m$ for clarity, while the actual base map used for matching is much larger ($500 m \times 500m$). We see that localization is quite reliable without the use of GNSS priors.

Most prior work in the field of localization ignores conditions such as night-time, and a nuanced evaluation in such subsets is important to build trustworthy AV subsystems. Furthermore, many other visual localization methods for AVs are trained on Google Street View data, which exclusively has day-time data. Hence, their failure modes at adverse operating conditions such as night-time are not as exposed. Comprehensive analysis across all such conditions will enable AVs to be safer for widespread deployment.
\vspace{-1mm}
\section{Conclusion}\label{sec:conclusion}
Reliable re-localization in GNSS-challenged environments such as urban canyons is an important yet challenging problem for autonomous driving. We proposed \setting{}, a method for GNSS-free re-localization that uses lidar and camera sensor data to find a highly accurate localization (often $< 1m$ accuracy) match within a neighborhood map as large as $500m\times500m$. \setting{} consists of a BEV map segmentation stage, followed by a coarse map patch retrieval method and finally a fine-grained matcher. We show that our method is robust across lighting conditions and improves in performance on using more generated frames across time. Moreover, we achieve a coarse localization accuracy of about $90\%$ for identifying the most-likely region, eventually resulting in an absolute recall$@1m$ of $39.8\%$ on a challenging NuScenes-derived dataset \textit{without} GNSS priors. \setting{} performs nearly twice as well as existing re-localization methods on the recall$@1m$ metric. \setting{} enables robust localization without GNSS priors in adverse operating conditions for AVs including GNSS-denied environments. 
\vspace{-1mm}
\printbibliography

@misc{wen20213dlidaraidedgnss,
      title={3D LiDAR Aided GNSS NLOS Mitigation in Urban Canyons}, 
      author={Weisong Wen and Li-Ta Hsu},
      year={2021},
      eprint={2112.06108},
      archivePrefix={arXiv},
      primaryClass={cs.RO},
}

@article{wang2013sequential,
  title={Sequential quadratic method for GPS NLOS positioning in urban canyon environments},
  author={He-Sheng Wang et al.},
  journal={International Journal of Automation and Smart Technology},
  volume={3},
  number={1},
  pages={37--46},
  year={2013}
}

@article{kaess_etal,
author = {Kaess, Michael and Johannsson, Hordur and Roberts, Richard and Ila, Viorela and Leonard, John J and Dellaert, Frank},
title = {iSAM2: Incremental smoothing and mapping using the Bayes tree},
year = {2012},
issue_date = {February  2012},
publisher = {Sage Publications, Inc.},
address = {USA},
volume = {31},
number = {2},
issn = {0278-3649},
doi = {10.1177/0278364911430419},
journal = {Int. J. Rob. Res.},
month = feb,
pages = {216–235},
numpages = {20}
}

@misc{zhou2020end,
      title={End-to-End Multi-View Fusion for 3D Object Detection in LiDAR Point Clouds}, 
      author={Yin Zhou and Pei Sun and Yu Zhang and Dragomir Anguelov and Jiyang Gao and Tom Ouyang and James Guo and Jiquan Ngiam and Vijay Vasudevan},
      year={2019},
      eprint={1910.06528},
      archivePrefix={arXiv},
      primaryClass={cs.CV}
}

@article{farag2022self,
  title={Self-driving vehicle localization using probabilistic maps and unscented-kalman filters},
  author={Farag, Wael},
  journal={International Journal of Intelligent Transportation Systems Research},
  volume={20},
  number={3},
  pages={623--638},
  year={2022},
  publisher={Springer}
}

@INPROCEEDINGS{8675596,
  author={Daniel Wilbers et al.},
  booktitle={2019 Third IEEE International Conference on Robotic Computing (IRC)}, 
  title={A Comparison of Particle Filter and Graph-Based Optimization for Localization with Landmarks in Automated Vehicles}, 
  year={2019},
  volume={},
  number={},
  pages={220-225},
  doi={10.1109/IRC.2019.00040}}

@ARTICLE{8910578,
  author={Perea-Strom, Daniel and Morell, Antonio and Toledo, Jonay and Acosta, Leopoldo},
  journal={IEEE Sensors Journal}, 
  title={GNSS Integration in the Localization System of an Autonomous Vehicle Based on Particle Weighting}, 
  year={2020},
  volume={20},
  number={6},
  pages={3314-3323},
  doi={10.1109/JSEN.2019.2955210}}

@ARTICLE{9833300,
  author={A Chalvatzaras et al.},
  journal={IEEE Transactions on Intelligent Vehicles}, 
  title={A Survey on Map-Based Localization Techniques for Autonomous Vehicles}, 
  year={2023},
  volume={8},
  number={2},
  pages={1574-1596},
  doi={10.1109/TIV.2022.3192102}}

@inproceedings{caesar2020nuscenes,
  title={nuscenes: A multimodal dataset for autonomous driving},
  author={Caesar, Holger and Bankiti, Varun and Lang, Alex H and Vora, Sourabh and Liong, Venice Erin and Xu, Qiang and Krishnan, Anush and Pan, Yu and Baldan, Giancarlo and Beijbom, Oscar},
  booktitle={Proceedings of the IEEE/CVF conference on computer vision and pattern recognition},
  pages={11621--11631},
  year={2020}
}

@misc{sun2021loftrdetectorfreelocalfeature,
      title={LoFTR: Detector-Free Local Feature Matching with Transformers}, 
      author={Jiaming Sun and Zehong Shen and Yuang Wang and Hujun Bao and Xiaowei Zhou},
      year={2021},
      eprint={2104.00680},
      archivePrefix={arXiv},
      primaryClass={cs.CV},
}

@misc{edstedt2023romarobustdensefeature,
      title={RoMa: Robust Dense Feature Matching}, 
      author={Johan Edstedt and Qiyu Sun and Georg Bökman and Mårten Wadenbäck and Michael Felsberg},
      year={2023},
      eprint={2305.15404},
      archivePrefix={arXiv},
      primaryClass={cs.CV},
}

@INPROCEEDINGS{6126544,
  author={Rublee, Ethan and Rabaud, Vincent and Konolige, Kurt and Bradski, Gary},
  booktitle={2011 International Conference on Computer Vision}, 
  title={ORB: An efficient alternative to SIFT or SURF}, 
  year={2011},
  volume={},
  number={},
  pages={2564-2571},
  doi={10.1109/ICCV.2011.6126544}}

@article{lowe2004distinctive,
  title={Distinctive image features from scale-invariant keypoints},
  author={Lowe, David G},
  journal={International journal of computer vision},
  volume={60},
  number={2},
  pages={91--110},
  year={2004},
  publisher={Springer}
}

@inproceedings{sarlin2023orienternet,
  title={Orienternet: Visual localization in 2d public maps with neural matching},
  author={Sarlin, Paul-Edouard and DeTone, Daniel and Yang, Tsun-Yi and Avetisyan, Armen and Straub, Julian and Malisiewicz, Tomasz and Bulo, Samuel Rota and Newcombe, Richard and Kontschieder, Peter and Balntas, Vasileios},
  booktitle={Proceedings of the IEEE/CVF Conference on Computer Vision and Pattern Recognition},
  pages={21632--21642},
  year={2023}
}

@INPROCEEDINGS{10802757,
  author={Wu, Hang and Zhang, Zhenghao and Lin, Siyuan and Mu, Xiangru and Zhao, Qiang and Yang, Ming and Qin, Tong},
  booktitle={2024 IEEE/RSJ International Conference on Intelligent Robots and Systems (IROS)}, 
  title={MapLocNet: Coarse-to-Fine Feature Registration for Visual Re-Localization in Navigation Maps}, 
  year={2024},
  volume={},
  number={},
  pages={13198-13205},
  doi={10.1109/IROS58592.2024.10802757}}

@INPROCEEDINGS{7989239,
  author={Kim, Dong-Ki and Walter, Matthew R.},
  booktitle={2017 IEEE International Conference on Robotics and Automation (ICRA)}, 
  title={Satellite image-based localization via learned embeddings}, 
  year={2017},
  volume={},
  number={},
  pages={2073-2080},
  doi={10.1109/ICRA.2017.7989239}}

@INPROCEEDINGS{extended_vins_mono,
  author={He, Mengwen and Rajkumar, Ragunathan Raj},
  booktitle={2021 IEEE/RSJ International Conference on Intelligent Robots and Systems (IROS)}, 
  title={Extended VINS-Mono: A Systematic Approach for Absolute and Relative Vehicle Localization in Large-Scale Outdoor Environments}, 
  year={2021},
  volume={},
  number={},
  pages={4861-4868},
  doi={10.1109/IROS51168.2021.9636776}}

@INPROCEEDINGS{7989618,
  author={Dubé, Renaud and Dugas, Daniel and Stumm, Elena and Nieto, Juan and Siegwart, Roland and Cadena, Cesar},
  booktitle={2017 IEEE International Conference on Robotics and Automation (ICRA)}, 
  title={SegMatch: Segment based place recognition in 3D point clouds}, 
  year={2017},
  volume={},
  number={},
  pages={5266-5272},
  doi={10.1109/ICRA.2017.7989618}}

@INPROCEEDINGS{7995900,
  author={Akai, Naoki and Morales, Luis Yoichi and Takeuchi, Eijiro and Yoshihara, Yuki and Ninomiya, Yoshiki},
  booktitle={2017 IEEE Intelligent Vehicles Symposium (IV)}, 
  title={Robust localization using 3D NDT scan matching with experimentally determined uncertainty and road marker matching}, 
  year={2017},
  volume={},
  number={},
  pages={1356-1363},
  doi={10.1109/IVS.2017.7995900}}

@misc{he2015deepresiduallearningimage,
      title={Deep Residual Learning for Image Recognition}, 
      author={Kaiming He and Xiangyu Zhang and Shaoqing Ren and Jian Sun},
      year={2015},
      eprint={1512.03385},
      archivePrefix={arXiv},
      primaryClass={cs.CV},
}

@misc{oquab2024dinov2learningrobustvisual,
      title={DINOv2: Learning Robust Visual Features without Supervision}, 
      author={Maxime Oquab and Timothée Darcet and Théo Moutakanni and Huy Vo and Marc Szafraniec and Vasil Khalidov and Pierre Fernandez and Daniel Haziza and Francisco Massa and Alaaeldin El-Nouby and Mahmoud Assran and Nicolas Ballas and Wojciech Galuba and Russell Howes and Po-Yao Huang and Shang-Wen Li and Ishan Misra and Michael Rabbat and Vasu Sharma and Gabriel Synnaeve and Hu Xu and Hervé Jegou and Julien Mairal and Patrick Labatut and Armand Joulin and Piotr Bojanowski},
      year={2024},
      eprint={2304.07193},
      archivePrefix={arXiv},
      primaryClass={cs.CV},
}

@inproceedings{camiletto2024u,
  title={U-bev: Height-aware bird’s-eye-view segmentation and neural map-based relocalization},
  author={Camiletto, Andrea Boscolo and Bochicchio, Alfredo and Liniger, Alexander and Dai, Dengxin and Gawel, Abel},
  booktitle={2024 IEEE/RSJ International Conference on Intelligent Robots and Systems (IROS)},
  pages={5597--5604},
  year={2024},
  organization={IEEE}
}

@inproceedings{wang2024efficient,
  title={Efficient LoFTR: Semi-dense local feature matching with sparse-like speed},
  author={Wang, Yifan and He, Xingyi and Peng, Sida and Tan, Dongli and Zhou, Xiaowei},
  booktitle={Proceedings of the IEEE/CVF conference on computer vision and pattern recognition},
  pages={21666--21675},
  year={2024}
}

@inproceedings{sural2024contextvlm,
  author={Shounak Sural et al.},
  booktitle={2024 IEEE 27th International Conference on Intelligent Transportation Systems (ITSC)}, 
  title={ContextVLM: Zero-Shot and Few-Shot Context Understanding for Autonomous Driving Using Vision Language Models}, 
  year={2024},
  volume={},
  number={},
  pages={468-475},
  doi={10.1109/ITSC58415.2024.10920066}}

@inproceedings{sural2024contextualfusion,
  title={ContextualFusion: Context-based multi-sensor fusion for 3D object detection in adverse operating conditions},
  author={Shounak Sural et al.},
  booktitle={IEEE intelligent vehicles symposium (IV)},
  pages={1534--1541},
  year={2024},
  organization={IEEE}
}

@article{liu2022bevfusion,
  title={Bevfusion: Multi-task multi-sensor fusion with unified bird's-eye view representation},
  author={Liu, Zhijian and Tang, Haotian and Amini, Alexander and Yang, Xinyu and Mao, Huizi and Rus, Daniela and Han, Song},
  journal={arXiv preprint arXiv:2205.13542},
  year={2022}
}

@inproceedings{zou2024m,
  title={M2 Depth: Self-supervised Two-Frame Multi-camera Metric Depth Estimation},
  author={Zou, Yingshuang and Ding, Yikang and Qiu, Xi and Wang, Haoqian and Zhang, Haotian},
  booktitle={European Conference on Computer Vision},
  pages={269--285},
  year={2024},
  organization={Springer}
}

@misc{hu2023planningorientedautonomousdriving,
      title={Planning-oriented Autonomous Driving}, 
      author={Yihan Hu and Jiazhi Yang and Li Chen and Keyu Li and Chonghao Sima and Xizhou Zhu and Siqi Chai and Senyao Du and Tianwei Lin and Wenhai Wang and Lewei Lu and Xiaosong Jia and Qiang Liu and Jifeng Dai and Yu Qiao and Hongyang Li},
      year={2023},
      eprint={2212.10156},
      archivePrefix={arXiv},
      primaryClass={cs.CV},
}

@article{he2025matchanything,
  title={Matchanything: Universal cross-modality image matching with large-scale pre-training},
  author={He, Xingyi and Yu, Hao and Peng, Sida and Tan, Dongli and Shen, Zehong and Bao, Hujun and Zhou, Xiaowei},
  journal={arXiv preprint arXiv:2501.07556},
  year={2025}
}

@misc{wang2023unitrunifiedefficientmultimodal,
      title={UniTR: A Unified and Efficient Multi-Modal Transformer for Bird's-Eye-View Representation}, 
      author={Haiyang Wang and Hao Tang and Shaoshuai Shi and Aoxue Li and Zhenguo Li and Bernt Schiele and Liwei Wang},
      year={2023},
      eprint={2308.07732},
      archivePrefix={arXiv},
      primaryClass={cs.CV},
}
\end{document}